\documentclass[10pt, a4paper]{article}
\usepackage{lrec2022} 
\usepackage{multibib}
\newcites{languageresource}{Language Resources}
\usepackage{graphicx}
\usepackage{tabularx}
\usepackage{soul}
\usepackage{parskip}
\usepackage{titlesec}
\titleformat{\section}{\normalfont\large\bfseries\center}{\thesection.}{1em}{}
\titleformat{\subsection}{\normalfont\SmallTitleFont\bfseries\raggedright}{\thesubsection.}{1em}{}
\titleformat{\subsubsection}{\normalfont\normalsize\bfseries\raggedright}{\thesubsubsection.}{1em}{}
\renewcommand\thesection{\arabic{section}}
\renewcommand\thesubsection{\thesection.\arabic{subsection}}
\renewcommand\thesubsubsection{\thesubsection.\arabic{subsubsection}}

\usepackage{epstopdf}
\usepackage[utf8]{inputenc}

\usepackage{hyperref}
\usepackage{xstring}

\usepackage{color}

\title{SimRelUz: Similarity and Relatedness scores as a Semantic Evaluation dataset for Uzbek language}

\name{Ulugbek Salaev$^{\ast}$, Elmurod Kuriyozov$^{\dagger}$, Carlos G\'omez-Rodr\'iguez$^{\dagger}$}

\address{$^{\ast}$Urgench State University, Department of Information                  Technologies\\
        14, Kh.Alimdjan str, Urgench city, 220100, Uzbekistan\\ 
        ulugbek0302@gmail.com \\ \\
        $^{\dagger}$Universidade da Coru\~na, CITIC\\ 
        Grupo LYS, Depto. de Computaci\'on y Tecnologías de la Información\\ 
        Facultade de Inform\'atica, Campus de Elvi\~na, A Coru\~na 15071, Spain\\ 
        \{e.kuriyozov, carlos.gomez\}@udc.es\\}

\abstract{
Semantic relatedness between words is one of the core concepts in natural language processing, thus making semantic evaluation an important task. In this paper, we present a semantic model evaluation dataset: SimRelUz - a collection of similarity and relatedness scores of word pairs for the low-resource Uzbek language. The dataset consists of more than a thousand pairs of words carefully selected based on their morphological features, occurrence frequency, semantic relation, as well as annotated by eleven native Uzbek speakers from different age groups and gender. We also paid attention to the problem of dealing with rare words and out-of-vocabulary words to thoroughly evaluate the robustness of semantic models. 
\\ 
\newline  }

\begin{document}

\maketitleabstract


\section{Introduction}

Having computational models that can measure the semantic relatedness and semantic similarity between concepts or words is an important fundamental task for many Natural Language Processing (NLP) applications, such as word sense disambiguation \cite{navigli2009word,agirre2007word}, thesauri, automatic dictionary generation \cite{mihalcea2001automatic,solovyev2020generation}, as well as machine translation \cite{bahdanau2014neural,brown1990statistical}. There are many language models that have been created that yield good quality semantic knowledge, yet their evaluation depends on gold standard datasets that have word/concept pairs scored by their semantic relations (such as synonymy, antonymy, meronymy, hypernymy, etc.), 
that come with cost due to their time-consuming context-generation process and high dependence on human annotators.

Many such datasets have been created so far for resource-rich languages \cite{hill2015simlex,finkelstein2001placing,rubenstein1965contextual}.
However, there is still a big gap of such datasets available for low-resource languages. Current work aims to fill that gap by providing, to our knowledge, the first semantic similarity and relatedness dataset for Uzbek language.
In this paper, we describe all the steps we followed as a set of data collection and annotation guidelines, with the full statistics and results obtained.
The main contributions of this paper are two-fold:
\begin{itemize}
\item Publicly available word pair semantic similarity and relatedness scoring web-based questionnaire software\footnote{Demo website: \url{https://simrel.urdu.uz}};
\item Publicly available semantic evaluation dataset including both similarity and relatedness scores for the low-resource Uzbek language \footnote{Both publicly available dataset and the source code of the web-application can be found here: \url{https://github.com/UlugbekSalaev/SimRelUz}.};
\end{itemize}
Furthermore, this paper also describes some important construction considerations about the dataset considering morphological and semantic attributes for a morphologically rich language, with their visualisations.


\textbf{Uzbek language}  (native: \textit{O‘zbek tili}) is a member of the Eastern Turkic or Karluk branch of the Turkic language family, an official language of Uzbekistan, and also a second language in neighbouring Central-Asian countries. It has more than 30 million speakers inside Uzbekistan alone, and more than ten million elsewhere in Central Asian countries, Southern Russian Federation, as well as the North-Eastern part of China, making it the second most widely spoken language among Turkic languages (right after Turkish language)\footnote{More information about Uzbek language: \url{https://en.wikipedia.org/wiki/Uzbek_language}}.

This paper has been organised as follows: It starts with a terminology section, explaining the basic definitions of terms used in the paper, then comes a related work section followed by a description of dataset creation and annotation process, moving onto some insights of the dataset, and in the end, authors describe their discussions, conclusions, as well as future work.

\section{Terminology}
In order to eliminate repetition, and to avoid confusion understanding the terms used in this paper, the terms similarity, relatedness, association, and distance may come with or without the prefix "semantic" interchangeably, but they are meant to mean the same respectively.

The term \textit{\textbf{semantic similarity}} in general, stands for a sense of relatedness that is dependent on the amount of shared properties, thus the 'degree of synonymy'. Whereas the term \textit{\textbf{semantic relatedness}} means a general sense of semantic proximity or semantic association, regardless of the causes of the connection humans can perceive. 
For instance \textit{bus/train} are good examples of semantic similarity, where they share many properties, i.e. they are both means of transport, both consume similar sorts of energy, have engines to operate, etc. On the other hand, \textit{teapot/cup} can be a good example of semantic relatedness, where they don't necessarily share common properties, but they are used in a similar context, since they both store tea, but teapot is for steeping tea in larger amounts, while a cup is for serving and drinking tea in smaller portions. Both above-mentioned examples can be used for semantic relatedness though, which means that semantic similarity is included inside semantic relatedness. Therefore, semantically similar things are, at the same time, semantically related, but the converse cannot be said to be the case in general.

\section{Related Work}
The first creation of a stand-alone semantic relation evaluation dataset dates back to the RG dataset~\cite{rubenstein1965contextual} , which was created for semantic similarity more than relatedness\footnote{RG dataset: \url{https://aclweb.org/aclwiki/RG-65_Test_Collection_(State_of_the_art)}}. Although it was very small in size (limited to only 65 noun pairs), it clearly showed the scientific importance, so the research interest continued later with more datasets coming along. The FrameNet \cite{baker1998berkeley} dataset is a rich linguistic resource with morphological, as well as expert-annotated semantic information as well. Among the most important gold-standard semantic evaluation datasets, we can find the WordSim-353 \cite{finkelstein2001placing}, MEN \cite{bruni2012distributional}, and SimLex-999 \cite{hill2015simlex} datasets for English. WordSim-353\footnote{ WordSim-353 datset: \url{http://alfonseca.org/eng/research/wordsim353.html}} contains 353 noun pairs scored by multiple human annotators. Similar to SimLex-353, the MEN\footnote{MEN dataset: \url{https://staff.fnwi.uva.nl/e.bruni/MEN}} dataset also is described as having similarity and relatedness distinctly, but the annotators only were asked to rate based on semantic relatedness. Later, introduction of the SimLex-999\footnote{SimLex-999 dataset: \url{https://fh295.github.io//simlex.html}} dataset made it the state-of-the-art gold standard semantic relatedness evaluation source. Some popular datasets for other languages include the RG dataset's German translation \cite{gurevych2005using}, the database of paradigmatic semantic relation pairs for German \cite{scheible2014database}, and the Simlex-999's translation into three languages: Italian, German and Russian  \cite{leviant2015judgment}. The Multi-SimLex \cite{vulic2020multi} project includes datasets for 12 diverse languages, including both major languages (English, Russian, Chinese, etc.) and less-resourced ones (Welsh, Kiswahili). Multi-SimLex\footnote{Multi-SimLex project and dataset: \url{https://multisimlex.com}} was a project originated from Simlex-999, and was taken to another step by creating a larger and more comprehensive dataset.  Linguistic databases such as VerbNet \cite{schuler2005verbnet} and WordNet \cite{miller1995wordnet,fellbaum2010wordnet} together with their implementations for other languages also contain semantically rich information created by experts. 

Since this is the first work of this kind for Uzbek language, the closest related work would be the related resources created for other Turkic languages, such as Turkish WordNets \cite{tufis2004balkanet,bakay2021turkish}, and especially AnlamVer dataset \cite{ercan2018anlamver}, where it contains both semantic similarity and relatedness scores annotated by many native speakers. Furthermore, the AnlamVer also shares useful knowledge of dataset design consideration when dealing with morphologially-rich and agglutinative languages.

\paragraph{Work on Uzbek language.}
Although there have been many papers published claiming that they have created NLP resources or developed some useful tools for Uzbek language, most of them, according to humble search results gathered by the authors, turned out to be ``zigglebottom'' papers \cite{pedersen2008last}. However, there are also many useful papers with  publicly available resources, some of them are the first Uzbek morphological analyzer \cite{matlatipov2009representation}, transliteration \cite{mansurov2021uzbek}, WordNet type synsets \cite{agostini2021uzwordnet}, Uzbek stopwords dataset \cite{Madatov_Bekchanov_Vicic_2021}, sentiment analysis \cite{rabbimov2020investigating,kuriyozov2019building}, text classification \cite{rabbimov2020multi}, and even a recent pretrained Uzbek language model based on the BERT architecture \cite{mansurov2021uzbert}.
There is also a well established Finite State Transducer(FST) based morphological analyzer for Uzbek language with more than 60K lexemes in Apertium monolingual package\footnote{\url{https://github.com/apertium/apertium-uzb}}.

\section{Dataset Design and Methodology}
The criterion for the construction of the dataset had to satisfy all the requirements available to make a high-quality semantic evaluation resource. So we followed the design choice and recommendations brought by authors of previous work \cite{finkelstein2001placing,bruni2012distributional,hill2015simlex,ercan2018anlamver,vulic2020multi}, such as follows:
\begin{itemize}
\item \verb|Clear definition|: The dataset must provide a clear definition of what semantic relation is supposed to be scored. So we decided to collect scores of both similarity and relatedness separately;
\item \verb|Language representativity|: The dataset should should be built considering diverse concepts of the language, such as parts of speech (i.e. verb, noun, adjective, ...), word formations (root, inflectional, or derivative), possible semantic relations (i.e. synonymy, antonymy, meronymy, ...), as well as the frequency range (i.e. frequent words, rare words, even out-of-vocabulary words);
\item \verb|Consistency and reliability|: Clear and precise scoring guidelines were provided to get consistent annotations from native speakers with different level of linguistic expertise.
\end{itemize}

More detailed information regarding each criteria are given below.

\subsection{Design choice}
\begin{figure*}[!ht]
    \centering
    \includegraphics[width=0.6\textwidth]{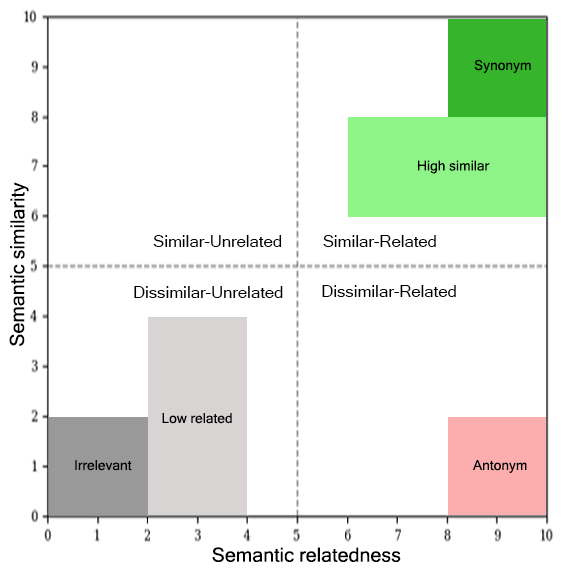}
    \caption{Semantic relation vector-space (proposed by AnlamVer project).}
    \label{fig:anlamver-sim-rel-space}
\end{figure*}

For the design of the dataset we followed the AnlamVer project \cite{ercan2018anlamver}, where instead of building two separate datasets for semantic similarity and relatedness, we decided to rate each word pair with two separate scores: one for similarity, and another for relatedness. This way, the resulting dataset was smaller in size, but richer in information. Moreover, this approach gave us an opportunity to visualize the dataset as a semantic relation space, using two scores as two dimensions, and creating a scatter plot. According to the methodology proposed by AnlamVer \cite{ercan2018anlamver} project, it is possible to predict the semantic relation of word pairs, by their location in the "Sim-Rel vector space", which is given in Figure \ref{fig:anlamver-sim-rel-space}. 

\subsection{Word candidates selection}
Probably a relatively easy way to obtain candidate words with minimum work would be translating words from gold-standard resources available for rich-resource languages (i.e. Multi-Simlex \cite{vulic2020multi}). However, there have been various relevant problems that have been reported to be caused by the use of such  translations, such as:
\begin{itemize}
    \item Two synonym pairs from a source language being mapped to one word in target language (Both words in \textit{car - automobile} pair in English would be mapped to a single \textit{avtomobil} in Uzbek);
    \item A translation of a single word in a source language that makes it multiple words in a target one (the word \textit{asylum} in English would be translated as \textit{ruhiy kasalliklar shifoxonasi} in Uzbek);
    \item Loss in the similarity/relatedness scores due to other cross-lingual aspects of pairs, such as translation accuracy or semantic/grammatical/cultural differences, require human annotators to re-score, leaving the costly part to be done again.
\end{itemize}
Therefore, we decided to choose the candidate word-list ourselves for better quality.
The first thing to make was a comprehensive list of words in the language using a big language corpus.
For the language corpus mentioned in this work, we used the Uzbek corpus from the CUNI corpora for Turkic languages \cite{baisa2012large}, which is, to our knowledge, the biggest Uzbek corpus collected with 18M tokens.
To obtain their part-of-speech (POS) tags, we used the UzWordNET dataset \cite{agostini2021uzwordnet} (which contains very limited information of root words with their POS classes), and Apertium-Uzb monolingual data\footnote{\url{https://github.com/apertium/apertium-uzb}} (contains more than 60K of Uzbek root words with their POS tags). Then we extracted nouns, adjectives and verbs only (with descending order relatively, according to their frequencies in the corpus), following the custom of similar gold-standard semantic evaluation resources.
Apart from only root forms of words, we also did manual selection of words with inflectional and derivational forms of words.

\subsection{Frequency-based considerations}
Considering the agglutinative nature of Uzbek language, creating the list of word frequencies in this language is not an easy task, since a single word can occur together with many different morphemes (either a single morpheme or a combination of many), making it difficult to obtain the actual count of occurrences of a single root-word. 
In this paper, we created a list of stems with their frequencies in Uzbek language using the biggest available Uzbek corpora \cite{baisa2012large}.
Firstly, the CUNI corpus was tokenized into sentences, then all the sentences were fed to the Apertium morphological analyser tool for Uzbek language\footnote{Although we have used the CLI version of the Apertium morphoological analyzer, it also can be accessed on the web to check its features: \url{https://turkic.apertium.org/index.eng.html?choice=uzb##analyzation}}.
Then, all the parts except for the lemmas of the resulting output were removed, which allowed us to obtain a stem/root-word frequency list.
Our priority was to include as many words with different frequencies as possible, so we used a technique similar to the one issued by the RareWords dataset \cite{luong2013better} - grouping words by their frequencies, 
dividing into three groups labeled as \textit{low}, \textit{medium}, \textit{high} with [2,5],[6,49],[50+] count ranges respectively.

\subsection{Rare and OOV words}
Furthermore, to make the dataset useful for checking the robustness of the semantic models, considering less-frequent words, even words that do not exist in the language dictionary but might appear in the context due to some morphological (surface words), syntactical (typo), or phonetical (homophones) reasons is also an important aspect.
Thus, the words where their root form does not appear  more than 3 times in the corpus were grouped as rare words, and their representatives were manually selected for the word list.\\
Considering the rich morphological aspect of Uzbek language, like other Turkic languages, there is a high inflection and derivation rate, where words are made in an agglutinative way: by combining stem and one or more morphemes (as prefix or suffixes). Hence, there is a high chance that a word may be grammatically wrong, but was created following surface-word creation rules (of which almost an unlimited number can be created). So we chose the following two most common out-of-vocabulary word cases, which are formally incorrect, but considered as acceptable forms for native speakers, and added some examples to the dataset:
\begin{itemize}
\item \verb|Stem-morpheme ambiguation|: It is a frequent case in Uzbek where stem and morpheme are combined directly, skipping the slight changes to fit them. E.g. \textit{yaxshiliq} instead of \textit{yaxshilik} (goodness), \textit{qamoqga} instead of \textit{qamoqqa} (to jail);
\item \verb|Phonetic ambiguation|: Two letters in Uzbek alphabet: ``x'' and ``h'' are phonetically so close to each-other, it is hard to identify them when used in a context, so people frequently mistake one for another when writing. E.g. \textit{pahta} instead of \textit{paxta} (cotton), \textit{shaxzoda} instead of \textit{shahzoda} (prince).
\end{itemize}
In total, 128 examples from both rare and OOV words with diverse POS types and word forms were added to the dataset.

After going through all the above mentioned steps and considerations, we gathered 1963 unique words to construct pairs. All their distribution among ford types, word forms, as well as word frequencies are given in Table \ref{table:unique-words-distribution}.

\begin{table*}[ht]
    \centering
    \begin{tabular}{lclclc}
        \multicolumn{2}{c|}{\textbf{Word classes}} & \multicolumn{2}{c|}{\textbf{Word forms}} & \multicolumn{2}{c}{\textbf{Word frequencies}} \\ \hline
        Nouns & \multicolumn{1}{c|}{1154} & Root form & \multicolumn{1}{c|}{995} & High frequency & 1136 \\
        Verbs & \multicolumn{1}{c|}{351} & Infelctional & \multicolumn{1}{c|}{423} & Medium frequency & 448 \\
        Adjectives & \multicolumn{1}{c|}{457} & Derivational & \multicolumn{1}{c|}{544} & Low frequency \& OOV & 378 \\ \hline
        \multicolumn{6}{c}{\textbf{Total number of unique words: 1962}}\\ \hline
    \end{tabular}
    \caption{Distribution of words by different word types, word forms, and word frequencies.}
    \label{table:unique-words-distribution}
\end{table*}

\begin{figure*}[ht]
    \centering
    \includegraphics[width=0.7\textwidth]{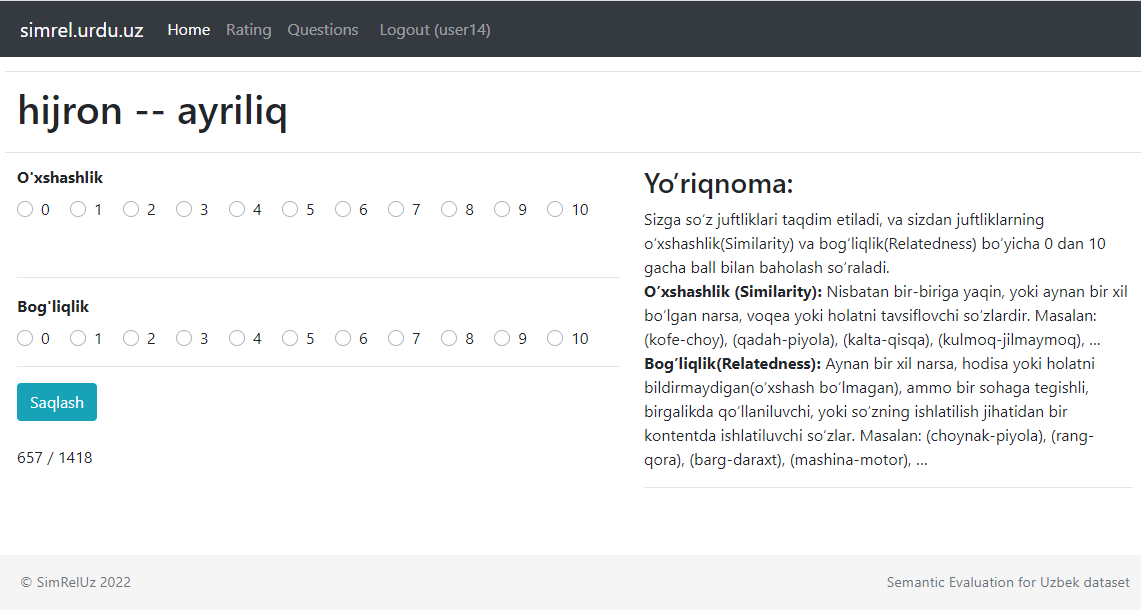}
    \caption{User interface of web-based annotation app.}
    \label{fig:app-user-interface}
\end{figure*}

\subsection{Word pairs selection}
Choosing word pairs randomly and scoring them would require the dataset to be huge in size, taking a very long time to annotate, so we tried to provide best quality semantic evaluation dataset with a limited number of word pairs by pre-establishing common semantic relations, such as synonymity, antonymity, hypernymity, and meronymity. This way the dataset would achieve a diverse distribution of scores, rather than filled up with very low scores due to most words not being related.
Thus, we selected common semantic relation categories, namely synonyms, antonyms, meronyms and hypernyms, and manually combined words from the word candidates list, tagging the pairs by a category where they most likely fit. Furthermore, we added word pairs by random allocation, which we named this category of pairs "irrelevant" (not in the sense of irrelevant pairs but in the sense of the magnitude of their semantic similarity and relatedness, as they are more likely to have very low scores on both sides). 

Overall, 1418 word pairs were selected for the annotation, Table \ref{table:word-pairs} shows the number of word pairs for each individual category.
\begin{table}[ht]
    \centering
    \begin{tabular}{lc}
        \textbf{Category} & \textbf{\# of word pairs} \\
        \hline
        Synonyms & 639 \\
        Antonyms & 239 \\
        Hypernyms & 220 \\
        Meronyms & 193 \\
        Irrelevant/Random & 127 \\
        \hline
        \textbf{Total} & \textbf{1418}
    \end{tabular}
    \caption{Distribution of word pairs by their pre-established semantic relations.}
    \label{table:word-pairs}
\end{table}

\section{Annotation process}
For the annotation process, we have created a web-based survey application where each annotator is given a unique username and password, where they can access the website and rate given word pairs with two separate scores at once. General user interface of the annotation page can be seen in Figure \ref{fig:app-user-interface}.

\begin{figure}[!ht]
    \centering
    \includegraphics[width=\columnwidth]{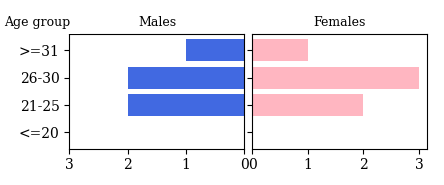}
    \caption{Distribution of annotators based on gender and age-groups.}
    \label{fig:age-gender-distribution}
\end{figure}

In total, eleven annotators (including two authors), who are native Uzbek speakers with different linguistic background, from different age groups and genders, have participated at the annotation, rating each pair once, with two scores (one for similarity, and the other for relatedness) from 0 to 10. Based on a statistical analysis from \cite{snow2008cheap}, more than ten annotators for a semantic evaluation are reliable enough.
In the end, there were eleven scores of similarity and the same amount for relatedness for each word pair, and we took their 
averages as the final scores. Figure \ref{fig:age-gender-distribution} shows the distribution of age and gender between annotators.

\begin{figure*}[ht]
    \centering
    \includegraphics[width=0.8\textwidth]{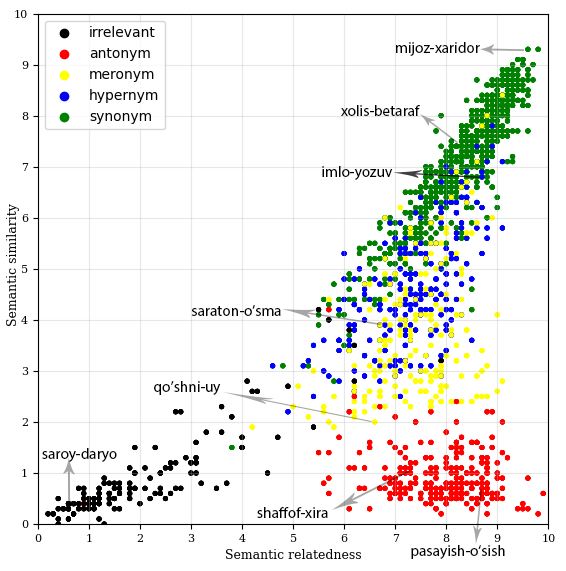}
    \caption{Visualisation of the created dataset in a Sim-Rel vector space.}
    \label{fig:result-scatter}
\end{figure*}

\section{Results}
The resulting dataset is composed of 1418 word pairs from different word types (nouns, adjectives and verbs), different word forms (root, inflectional, derivational), with different frequencies (high, mid, low frequencies, rare and OOV words), and with diverse pre-established semantic relations (synonym, antonym, meronym, hypernym, not related). All the pairs have two scores, one for semantic similarity, while the other is for semantic relatedness. No field in the dataset was left empty (as was requested from annotators in the guidelines, even for the OOV cases), and the average pairwise inter-annotator agreement scores (apia) were computed for both semantic similarity and relatedness separately, where we achieved 0.71 and 0.69 apia scores for semantic similarity and relatedness respectively, meaning that although we have scored less than AnlamVer dataset (0.75), it still performed better than most semantic evaluation datasets (SimLex=0.67, MEN=0.68).
The resulting dataset can be plotted into the Sim-Rel vector space as shown in Figure \ref{fig:result-scatter}.

\paragraph{Discussions.}
As can be seen from the scatter plot of the dataset in a vector space (Figure \ref{fig:result-scatter}), it can be concluded that average scores of word pairs visually correlate to our pre-established relation types, since they are scattered mostly inside and around the determined areas in the vector-space. Irrelevant and random pairs can be easily detected from the plot, that it has no much overlap with other types. It is also worth mentioning that none of the word pair is in the Similar-Unrelated (top-left quarter of the vector-space) part of the plot, confirming its reliability, since a word cannot be similar, but not related at once. There is a big overlap between hypernym, meronym, and partially synonym pairs, as expected, as they share similar score ranges. 
Handling OOV words by annotators has also met our expectations, where they treated them as regular words and scored accordingly.


\section{Conclusion}
In this paper, we presented SimRelUz, a novel semantic evaluation dataset for the low-resource Uzbek language, with semantic similarity and relatedness scores for 1418 word pairs, which were selected based on their morphological classes, word-forms, frequencies, also including rare and out-of-vocabulary words for better evaluation of semantic language models.
This kind of dataset is a useful resource to be used for evaluation of computational semantic analysis systems that will be created in the future for Uzbek, in simpler words, for formal analysis of meaning in language models.
Moreover, we have also presented an open-source web-based semantic evaluation tool designed for multiple-user annotation. 
Our future work includes intrinsic and extrinsic analysis of created dataset, also creating big WordNet-type knowledge-base for Uzbek language. 


\section{Acknowledgements}
This work has received funding from ERDF/MICINN-AEI (SCANNER-UDC, PID2020-113230RB-C21), from Xunta de Galicia (ED431C 2020/11), and from Centro de Investigación de Galicia ``CITIC'', funded by Xunta de Galicia and the European Union (ERDF - Galicia 2014-2020 Program), by grant ED431G 2019/01. Elmurod Kuriyozov was funded for his PhD by El-Yurt-Umidi Foundation under the Cabinet of Ministers of the Republic of Uzbekistan. The authors would also like to thank the NLP team of Urgench State University for their tremendous help with the web hosting, and annotation.

\section{Bibliographical References}\label{reference}

\bibliographystyle{lrec2022-bib}
\bibliography{main}

\begin{thebibliography}{}

\bibitem[\protect\citename{Agirre and Edmonds}2007]{agirre2007word}
Agirre, E. and Edmonds, P.
\newblock (2007).
\newblock {\em Word sense disambiguation: Algorithms and applications},
  volume~33.
\newblock Springer Science \& Business Media.

\bibitem[\protect\citename{Agostini \bgroup et al.\egroup
  }2021]{agostini2021uzwordnet}
Agostini, A., Usmanov, T., Khamdamov, U., Abdurakhmonova, N., and Mamasaidov,
  M.
\newblock (2021).
\newblock Uzwordnet: A lexical-semantic database for the uzbek language.
\newblock In {\em Proceedings of the 11th Global Wordnet conference}, pages
  8--19.

\bibitem[\protect\citename{Bahdanau \bgroup et al.\egroup
  }2014]{bahdanau2014neural}
Bahdanau, D., Cho, K., and Bengio, Y.
\newblock (2014).
\newblock Neural machine translation by jointly learning to align and
  translate.
\newblock {\em arXiv preprint arXiv:1409.0473}.

\bibitem[\protect\citename{Baisa \bgroup et al.\egroup }2012]{baisa2012large}
Baisa, V., Suchomel, V., et~al.
\newblock (2012).
\newblock Large corpora for turkic languages and unsupervised morphological
  analysis.
\newblock In {\em Proceedings of the Eighth conference on International
  Language Resources and Evaluation (LREC’12), Istanbul, Turkey. European
  Language Resources Association (ELRA)}.

\bibitem[\protect\citename{Bakay \bgroup et al.\egroup }2021]{bakay2021turkish}
Bakay, {\"O}., Ergelen, {\"O}., Sarm{\i}{\c{s}}, E., Y{\i}ld{\i}r{\i}m, S.,
  Ar{\i}can, B.~N., Kocabalc{\i}o{\u{g}}lu, A., {\"O}z{\c{c}}elik, M.,
  San{\i}yar, E., Kuyruk{\c{c}}u, O., Avar, B., et~al.
\newblock (2021).
\newblock Turkish wordnet kenet.
\newblock In {\em Proceedings of the 11th global wordnet conference}, pages
  166--174.

\bibitem[\protect\citename{Baker \bgroup et al.\egroup
  }1998]{baker1998berkeley}
Baker, C.~F., Fillmore, C.~J., and Lowe, J.~B.
\newblock (1998).
\newblock The berkeley framenet project.
\newblock In {\em COLING 1998 Volume 1: The 17th International Conference on
  Computational Linguistics}.

\bibitem[\protect\citename{Brown \bgroup et al.\egroup
  }1990]{brown1990statistical}
Brown, P.~F., Cocke, J., Della~Pietra, S.~A., Della~Pietra, V.~J., Jelinek, F.,
  Lafferty, J., Mercer, R.~L., and Roossin, P.~S.
\newblock (1990).
\newblock A statistical approach to machine translation.
\newblock {\em Computational linguistics}, 16(2):79--85.

\bibitem[\protect\citename{Bruni \bgroup et al.\egroup
  }2012]{bruni2012distributional}
Bruni, E., Boleda, G., Baroni, M., and Tran, N.-K.
\newblock (2012).
\newblock Distributional semantics in technicolor.
\newblock In {\em Proceedings of the 50th Annual Meeting of the Association for
  Computational Linguistics (Volume 1: Long Papers)}, pages 136--145.

\bibitem[\protect\citename{Ercan and Y{\i}ld{\i}z}2018]{ercan2018anlamver}
Ercan, G. and Y{\i}ld{\i}z, O.~T.
\newblock (2018).
\newblock Anlamver: Semantic model evaluation dataset for turkish-word
  similarity and relatedness.
\newblock In {\em Proceedings of the 27th International Conference on
  Computational Linguistics}, pages 3819--3836.

\bibitem[\protect\citename{Fellbaum}2010]{fellbaum2010wordnet}
Fellbaum, C.
\newblock (2010).
\newblock Wordnet.
\newblock In {\em Theory and applications of ontology: computer applications},
  pages 231--243. Springer.

\bibitem[\protect\citename{Finkelstein \bgroup et al.\egroup
  }2001]{finkelstein2001placing}
Finkelstein, L., Gabrilovich, E., Matias, Y., Rivlin, E., Solan, Z., Wolfman,
  G., and Ruppin, E.
\newblock (2001).
\newblock Placing search in context: The concept revisited.
\newblock In {\em Proceedings of the 10th international conference on World
  Wide Web}, pages 406--414.

\bibitem[\protect\citename{Gurevych}2005]{gurevych2005using}
Gurevych, I.
\newblock (2005).
\newblock Using the structure of a conceptual network in computing semantic
  relatedness.
\newblock In {\em International conference on natural language processing},
  pages 767--778. Springer.

\bibitem[\protect\citename{Hill \bgroup et al.\egroup }2015]{hill2015simlex}
Hill, F., Reichart, R., and Korhonen, A.
\newblock (2015).
\newblock Simlex-999: Evaluating semantic models with (genuine) similarity
  estimation.
\newblock {\em Computational Linguistics}, 41(4):665--695.

\bibitem[\protect\citename{Kuriyozov and
  Matlatipov}2019]{kuriyozov2019building}
Kuriyozov, E. and Matlatipov, S.
\newblock (2019).
\newblock Building a new sentiment analysis dataset for uzbek language and
  creating baseline models.
\newblock In {\em Multidisciplinary Digital Publishing Institute Proceedings},
  volume~21, page~37.

\bibitem[\protect\citename{Leviant and Reichart}2015]{leviant2015judgment}
Leviant, I. and Reichart, R.
\newblock (2015).
\newblock Judgment language matters: Multilingual vector space models for
  judgment language aware lexical semantics.
\newblock {\em CoRR, abs/1508.00106}.

\bibitem[\protect\citename{Luong \bgroup et al.\egroup }2013]{luong2013better}
Luong, M.-T., Socher, R., and Manning, C.~D.
\newblock (2013).
\newblock Better word representations with recursive neural networks for
  morphology.
\newblock In {\em Proceedings of the seventeenth conference on computational
  natural language learning}, pages 104--113.

\bibitem[\protect\citename{Madatov \bgroup et al.\egroup
  }2021]{Madatov_Bekchanov_Vicic_2021}
Madatov, K., Bekchanov, S., and Vi{\v{c}}i{\v{c}}, J.
\newblock (2021).
\newblock Lists of uzbek stopwords.

\bibitem[\protect\citename{Mansurov and Mansurov}2021a]{mansurov2021uzbek}
Mansurov, B. and Mansurov, A.
\newblock (2021a).
\newblock Uzbek cyrillic-latin-cyrillic machine transliteration.
\newblock {\em arXiv preprint arXiv:2101.05162}.

\bibitem[\protect\citename{Mansurov and Mansurov}2021b]{mansurov2021uzbert}
Mansurov, B. and Mansurov, A.
\newblock (2021b).
\newblock Uzbert: pretraining a bert model for uzbek.
\newblock {\em arXiv preprint arXiv:2108.09814}.

\bibitem[\protect\citename{Matlatipov and
  Vetulani}2009]{matlatipov2009representation}
Matlatipov, G. and Vetulani, Z.
\newblock (2009).
\newblock Representation of uzbek morphology in prolog.
\newblock In {\em Aspects of Natural Language Processing}, pages 83--110.
  Springer.

\bibitem[\protect\citename{Mihalcea and Moldovan}2001]{mihalcea2001automatic}
Mihalcea, R. and Moldovan, D.~I.
\newblock (2001).
\newblock Automatic generation of a coarse grained wordnet.

\bibitem[\protect\citename{Miller}1995]{miller1995wordnet}
Miller, G.~A.
\newblock (1995).
\newblock Wordnet: a lexical database for english.
\newblock {\em Communications of the ACM}, 38(11):39--41.

\bibitem[\protect\citename{Navigli}2009]{navigli2009word}
Navigli, R.
\newblock (2009).
\newblock Word sense disambiguation: A survey.
\newblock {\em ACM computing surveys (CSUR)}, 41(2):1--69.

\bibitem[\protect\citename{Pedersen}2008]{pedersen2008last}
Pedersen, T.
\newblock (2008).
\newblock Last words: Empiricism is not a matter of faith.
\newblock {\em Computational Linguistics}, 34(3):465--470.

\bibitem[\protect\citename{Rabbimov and Kobilov}2020]{rabbimov2020multi}
Rabbimov, I. and Kobilov, S.
\newblock (2020).
\newblock Multi-class text classification of uzbek news articles using machine
  learning.
\newblock In {\em Journal of Physics: Conference Series}, volume 1546, page
  012097. IOP Publishing.

\bibitem[\protect\citename{Rabbimov \bgroup et al.\egroup
  }2020]{rabbimov2020investigating}
Rabbimov, I., Mporas, I., Simaki, V., and Kobilov, S.
\newblock (2020).
\newblock Investigating the effect of emoji in opinion classification of uzbek
  movie review comments.
\newblock In {\em International Conference on Speech and Computer}, pages
  435--445. Springer.

\bibitem[\protect\citename{Rubenstein and
  Goodenough}1965]{rubenstein1965contextual}
Rubenstein, H. and Goodenough, J.~B.
\newblock (1965).
\newblock Contextual correlates of synonymy.
\newblock {\em Communications of the ACM}, 8(10):627--633.

\bibitem[\protect\citename{Scheible and Im~Walde}2014]{scheible2014database}
Scheible, S. and Im~Walde, S.~S.
\newblock (2014).
\newblock A database of paradigmatic semantic relation pairs for german nouns,
  verbs, and adjectives.
\newblock In {\em Proceedings of Workshop on Lexical and Grammatical Resources
  for Language Processing}, pages 111--119.

\bibitem[\protect\citename{Schuler}2005]{schuler2005verbnet}
Schuler, K.~K.
\newblock (2005).
\newblock {\em VerbNet: A broad-coverage, comprehensive verb lexicon}.
\newblock University of Pennsylvania.

\bibitem[\protect\citename{Snow \bgroup et al.\egroup }2008]{snow2008cheap}
Snow, R., O’connor, B., Jurafsky, D., and Ng, A.~Y.
\newblock (2008).
\newblock Cheap and fast--but is it good? evaluating non-expert annotations for
  natural language tasks.
\newblock In {\em Proceedings of the 2008 conference on empirical methods in
  natural language processing}, pages 254--263.

\bibitem[\protect\citename{Solovyev \bgroup et al.\egroup
  }2020]{solovyev2020generation}
Solovyev, V., Bochkarev, V., and Khristoforov, S.
\newblock (2020).
\newblock Generation of a dictionary of abstract/concrete words by a multilayer
  neural network.
\newblock In {\em Journal of Physics: Conference Series}, volume 1680, page
  012046. IOP Publishing.

\bibitem[\protect\citename{Tufis \bgroup et al.\egroup
  }2004]{tufis2004balkanet}
Tufis, D., Cristea, D., and Stamou, S.
\newblock (2004).
\newblock Balkanet: Aims, methods, results and perspectives. a general
  overview.
\newblock {\em Romanian Journal of Information science and technology},
  7(1-2):9--43.

\bibitem[\protect\citename{Vuli{\'c} \bgroup et al.\egroup
  }2020]{vulic2020multi}
Vuli{\'c}, I., Baker, S., Ponti, E.~M., Petti, U., Leviant, I., Wing, K.,
  Majewska, O., Bar, E., Malone, M., Poibeau, T., et~al.
\newblock (2020).
\newblock Multi-simlex: A large-scale evaluation of multilingual and
  crosslingual lexical semantic similarity.
\newblock {\em Computational Linguistics}, 46(4):847--897.

\end{thebibliography}


\end{document}